%% file: main.tex
\documentclass[letterpaper]{article} 
\usepackage{aaai2026}  
\usepackage{times}  
\usepackage{helvet}  
\usepackage{courier}  
\usepackage[hyphens]{url}  
\usepackage{graphicx} 
\usepackage{natbib}  
\usepackage{caption} 
\usepackage{algorithm}
\usepackage{algpseudocode} 
\usepackage{newfloat}
\usepackage{listings}
\DeclareCaptionStyle{ruled}{labelfont=normalfont,labelsep=colon,strut=off} 
\floatstyle{ruled}
\newfloat{listing}{tb}{lst}{}
\floatname{listing}{Listing}
\usepackage{amsmath}
\usepackage{amsfonts}
\usepackage{booktabs}  
\usepackage{multirow}  
\usepackage{pifont}  
\usepackage{colortbl}  
\usepackage{xcolor}  
\usepackage{threeparttable}  
\usepackage{makecell}  
\usepackage{tabularx}  

 \usepackage{subcaption}

\title{MedSpaformer: a Transferable Transformer with Multi-granularity Token Sparsification for Medical Time Series Classification}

\author{
    \textsuperscript{\rm 1}Jiexia Ye,
    \textsuperscript{\rm 2}Weiqi Zhang,
    \textsuperscript{\rm 3}Ziyue Li,
    \textsuperscript{\rm 2}Jia Li,
    \textsuperscript{\rm 2}Fugee Tsung
}

\affiliations{
    \textsuperscript{\rm 1}The Hong Kong University of Science and Technology (Guangzhou), China\\
    jye324@connect.hkust-gz.edu.cn
    
    \textsuperscript{\rm 2}The Hong Kong University of Science and Technology, Hong Kong SAR, China\\
    weiqizhang@ust.hk, jialee@ust.hk, fgtsung@ust.hk
    
    \textsuperscript{\rm 3}University of Cologne, Germany\\
    zlibn@wiso.uni-koeln.de
}

\begin{document}

\maketitle

\begin{abstract}
Accurate medical time series (MedTS) classification is essential for effective clinical diagnosis, yet remains challenging due to complex multi-channel temporal dependencies, information redundancy, and label scarcity.
While transformer-based models have shown promise in time series analysis, most are designed for forecasting tasks and fail to fully exploit the unique characteristics of MedTS.
In this paper, we introduce MedSpaformer, a transformer-based framework tailored for MedTS classification. It incorporates a sparse token-based dual-attention mechanism that enables global context modeling and token sparsification, allowing dynamic feature refinement by focusing on informative tokens while reducing redundancy.
This mechanism is integrated into a multi-granularity cross-channel encoding scheme to capture intra- and inter-granularity temporal dependencies and inter-channel correlations, enabling progressive refinement of task-relevant patterns in medical signals.
The sparsification design allows our model to flexibly accommodate inputs with variable lengths and channel dimensions. We also introduce an adaptive label encoder to extract label semantics and address cross-dataset label space misalignment. Together, these components enhance the model’s transferability across heterogeneous medical datasets, which helps alleviate the challenge of label scarcity.
Our model outperforms 13 baselines across 7 medical datasets under supervised learning. It also excels in few-shot learning and demonstrates zero-shot capability in both in-domain and cross-domain diagnostics.
These results highlight MedSpaformer's robustness and its potential as a unified solution for MedTS classification across diverse settings.
\textbf{The code is provided in the supplementary material.}

\end{abstract}
\section{Introduction}
Medical time series (MedTS) data—such as multi-channel electrocardiograms (ECGs) and electroencephalograms (EEGs)—encode rich temporal dynamics crucial for diagnosing life-threatening conditions like arrhythmias \cite{wagner2020ptb} and epilepsy \cite{shah2018temple}. Early and accurate classification of these signals enables timely intervention and personalized treatment \cite{ DWangCHYSSJKSYRS22}.
Yet, MedTS data present unique modeling challenges due to their complex structure and clinical constraints:
First, MedTS signals exhibit \textit{complex multi-channel temporal dependencies}. Pathological patterns span diverse time scales—from millisecond-level epileptic spikes to minute-level slow oscillations—and are distributed across multiple sensors (e.g., 19-lead EEGs), requiring simultaneous modeling of both temporal hierarchy and cross-channel interactions \cite{wang2024medformer, Tang2021SelfSupervisedGN}.
Second, MedTS data are often \textit{redundant and noisy}, with repeated or irrelevant segments that dilute discriminative patterns and increase computational overhead \cite{ZHANG2024120239}.
Third, \textit{label scarcity} is pervasive—clinically annotated datasets are limited due to the high cost of expert labeling, particularly for rare disorders \cite{YangLWZYYZ25, li2024frozen}.

Traditional MedTS approaches rely on shallow statistical features \cite{rahman2015combining, riaz2020gaussian}. Current deep learning models—including RNNs \cite{Salloum_Kuo_2017}, CNNs \cite{Lawhern_2018}, and GNNs \cite{Tang2021SelfSupervisedGN}—are capable of capturing increasingly complex patterns.
Recently, transformer-based models have emerged as powerful sequence learners, particularly in time series forecasting \cite{wen2022transformers}. However, most of them are not specifically designed for MedTS classification, and thus fall short in addressing its domain-specific challenges.
PatchTST \cite{nie2023time} and Crossformer \cite{zhang2023crossformer} capture local patterns through patching but lack multi-scale flexibility due to fixed patch sizes.
MTST \cite{zhang2024multi} and Pathformer \cite{chen2024pathformer} introduce multi-resolution strategies but are confined to single-channel inputs.
In contrast, FEDformer \cite{zhou2022fedformer} and Autoformer \cite{wu2021autoformer} enable cross-channel attention but overlook multi-scale structure. Medformer \cite{wang2024medformer} unifies these views via multi-granularity cross-channel modeling, but its dense self-attention indiscriminately attends to all tokens, lacking effective suppression of redundant signals.
Furthermore, these models' rigid architectural designs—fixed input lengths and channel configurations—constrain their adaptability to heterogeneous datasets, hindering their potential to mitigate label scarcity through cross-dataset transfer learning.

To bridge these limitations, we propose MedSpaformer, a transferable transformer specifically designed for MedTS classification. First, we design a \textbf{T}oken-\textbf{S}parse \textbf{D}ual \textbf{A}ttention (TSDA) mechanism for granularity and channel modeling. TSDA employs self-attention to model global token interactions, followed by token-sparse attention to compress tokens using a fixed smaller number of domain-guided learnable queries. This sparsification aims to remove redundant information, preserve task-relevant features, and reduce computational cost. 
We stack multiple TSDA blocks to progressively encode multi-granularity and cross-channel information. They first capture local granular features, then refine inter-granularity dependencies, and finally model cross-channel interactions to integrate complementary information.
The sparse encoding of TSDA can transform input sequences of varying lengths into fixed-length, enabling our model to directly process heterogeneous inputs with different sequence length and channel configurations. Further, we design an adaptive label encoder to project label descriptions into a unified latent space to bridge cross-dataset label space mismatches. Together, they allow MedSpaformer to transfer knowledge across datasets with varying lengths, channels, and classes, demonstrating few-shot and zero-shot transferability in diverse clinical applications, mitigating the limited label challenge.
Our main contributions are as follows:

\begin{itemize}
\item
We propose MedSpaformer, a transformer architecture tailored for MedTS classification. By incorporating a token-sparse dual-attention mechanism into a multi-granularity cross-channel encoding framework, MedSpaformer progressively distills informative patterns, reduces redundancy, and effectively models multi-scale temporal dynamics and cross-channel dependencies in medical data.

\item
MedSpaformer supports input-output heterogeneity via the sparse encoding mechanism and an adaptive label encoder. To the best of our knowledge, it is the first transformer framework enabling cross-task zero-shot transfer in time series classification.
\item
We conduct extensive experiments on multiple public datasets, achieving state-of-the-art performance in both supervised and few-shot settings. Furthermore, we evaluate our model in in-domain and cross-domain zero-shot scenarios to demonstrate its cross-dataset transferability.

\end{itemize}

\section{Related Work}
\textbf{Medical Time Series Classification.}
Medical time series (MedTS) data, such as EEG \cite{escudero2006analysis}, ECG \cite{physiobank2000physionet}, EMG \cite{Xiong_2021}, and EOG \cite{Fan_2021}, are widely used in disease diagnosis, monitoring, and rehabilitation \cite{fatourechi2007emg}.
Traditional methods, such as nearest neighbor classifiers \cite{rahman2015combining}, auto-regressive models \cite{Schaffer_2021}, and Gaussian mixture models \cite{Vincent_2009}, offer simplicity and interpretability but face challenges when dealing with complex, high-dimensional patterns. With the advent of deep learning, models leveraging RNNs \cite{Salloum_Kuo_2017}, CNNs \cite{Lawhern_2018}, and GNNs \cite{Tang2021SelfSupervisedGN} have dominated MedTS classification. For instance,  EEGNet \cite{Lawhern_2018} uses depthwise separable convolutions to extract EEG features, while GNNs \cite{Tang2021SelfSupervisedGN} enable self-supervised seizure detection. While these models show promising results in tasks with single-modality medical signals, they often lack generalizability across different medical modalities. For example, a model tailored for ECG \cite{ding2025advances} may not transfer effectively to other types of medical signals such as EEG \cite{sharma2024emerging} or EOG \cite{van2024single}.

\textbf{Transformer for Time Series.}
Transformers have significantly advanced time series analysis. Based on tokenization strategies, they can be categorized into single-timestamp \cite{wu2021autoformer,zhou2021informer}, all-timestamp \cite{LiuHZWWML24}, and multi-timestamp approaches \cite{zhang2023crossformer,zhang2024multi,wang2024medformer}, with the latter further divided into single- and multi-granularity methods.
Single-timestamp tokenization struggles with capturing coarse-grained patterns, while all-timestamp strategies may overlook fine-grained local details. Single-granularity methods like PatchTST \cite{nie2023time} and Crossformer \cite{zhang2023crossformer} generate fixed-length patches from single-channel sequences, capturing local patterns but falling short in handling multi-scale dynamics.
Multi-granularity models such as MTST \cite{zhang2024multi} and Pathformer \cite{chen2024pathformer} address this by using varied patch sizes, yet remain limited to single-channel inputs, which may hinder performance for multivariate time series classification.
Medformer \cite{wang2024medformer} employs multi-granularity encoding and captures low-level channel correlations via cross-channel patching. In contrast, our model derives high-level channel representations through channel-wise multi-granularity encoding. Moreover, Medformer lacks a mechanism for suppressing redundant signals, which we address via token sparsification. Finally, while prior models show limited cross-dataset transferability, our model enables direct transfer across heterogeneous medical datasets.

 
\input{./Figures/tex/model}
\section{Methodology}
\textbf{Problem Formulation}
Consider a medical time series dataset $\mathcal{D}=\{(\mathbf{X}_i, y_i)\}_{i=1}^N$ where each signal $\mathbf{X}_i \in \mathbb{R}^{L \times C} $ contains $L$ timestamps across $C$ channels and each label $y_i \in \{1, 2, \ldots, M\}$ is described by a text $\mathcal{T}_{y_i}$. $M$ is the number of classes.
Our objective is to learn a framework to align the temporal signal $\mathbf{X}_i$ and its label description $\mathcal{T}_{y_i}$ into a unified ${D}$ dimension latent space to obtain their representations $\mathbf{h}_i^{(x)} \in \mathbb{R}^{D}$ and $\mathbf{h}_i^{(y)}\in \mathbb{R}^{D}$. The framework is optimized by maximizing the similarity between time series-label pairs $(\mathbf{h}_i^{(x)}, \mathbf{h}_i^{(y)})$.

\textbf{Overview}
 Figure \ref{fig:model} demonstrates our model. In this section, we first introduce the core component—the token-sparse dual attention block (TSDA). TSDA effectively captures global context among tokens, eliminates redundant signals and refines features by token sparsification. Next, we apply TSDA blocks on multi-granularity encoding for intra- and inter-granularity correlation extraction, and on multi-channel encoding for channel correlation integration. Built upon TSDA blocks, our model is inherently agnostic to input length and channel configurations.We also introduce an adaptive label encoder to align heterogeneous label spaces across datasets. These designs enable the model to be trained across diverse datasets and equip it with few-shot/zero-shot transferability across different medical applications.

\subsection{Token-Sparse Dual Attention Block}
Inspired by physicians’ two-stage diagnostic process—first holistically contextualizing symptoms, then analyzing specific biomarkers \cite{hausmann2016tracing}—we propose the Token-Sparse Dual Attention (TSDA) block, which mirrors this process through global context modeling and dynamic feature refinement using a two-stage attention mechanism. 

TSDA first employs self-attention to capture global long-range temporal dependencies, leveraging its proven ability to model pairwise token interactions in sequential data \cite{chen2024pathformer,wang2024medformer}. This global modeling capability integrates information across the entire sequence, reinforces inter-token dependencies, and contextualizes local patterns—crucial in medical signals, where waveform anomalies gain meaning only within the broader temporal structure \cite{wagner2020ptb}.
Formally, given an input sequence $\mathbf{H} \in \mathbb{R}^{L \times D}$, the self-attention output is defined as $\mathbf{H}^{\text{self}} \leftarrow \text{Attn}^{\text{self}}(\mathbf{H}, \mathbf{H}, \mathbf{H})$, where $\mathbf{H}^{\text{self}} \in \mathbb{R}^{L \times D}$.

While self-attention captures comprehensive temporal dependencies, medical signals often contain redundant or noisy patterns that obscure diagnostic features. To address this, inspired by Q-Former \cite{0008LSH23} which employs learnable queries to extract visual representation most relevant to the text, we design a token-sparse attention layer that uses learnable queries informed by domain-specific priors to selectively attend to diagnostically salient features—analogous to how physicians narrow their analysis to specific biomarkers with domain knowledge after forming an initial clinical impression.
Specifically, we introduce a set of $Q$ randomly initialized learnable query vectors $\mathbf{Q}$ , augmented with domain-specific prior embedding $\mathbf{e}^{\text {prior}}$: $\mathbf{Q}^{\text {aug}}=f(\mathbf{Q}, \mathbf{e}^{\text {prior}})$
where $\mathbf{Q}^{\text {aug}} \in \mathbb{R}^{Q \times D}$. $f$ is the function to fuse queries and priors and concatenation is applied in our experiments. Following \cite{jin2023timellm}, we utilize a frozen language model to generate domain-specific embedding based on the dataset description: $\mathbf{e}^{\text {prior}}=f_{\mathrm{LM}}(\mathcal{T}^\text{data})$. These queries then attend to $\mathbf{H}^{\text {self}}$ to generate a sparse token set:
\begin{equation}
\small
\begin{split}
  \mathbf{H}^{\text{sparse}} 
    &\leftarrow \text{Attn}^{\text{sparse}}(\mathbf{Q}^{\text{aug}}, \mathbf{H}^{\text{self}}, \mathbf{H}^{\text{self}}) \\
    &= \operatorname{Softmax}\left( \frac{(\mathbf{Q}^{\text{aug}}\mathbf{W}_Q) \left( \mathbf{H}^{\text{self}} \mathbf{W}_K \right)^{\top} }{ \sqrt{D} } \right) (\mathbf{H}^{\text{self}} \mathbf{W}_V)
\end{split}
\end{equation}
where $\mathbf{H}^{\text {sparse}} \in \mathbb{R}^{Q \times D}$ retains $Q$ tokens and $Q \ll L$, aiming to preserve critical features and eliminate irrelevant information, reducing computation.
Note that TSDA block transforms variable-length sequences into fixed-length representations via token-sparse attention. This design is input-length-agnostic—its trainable parameters depend solely on the predefined queries number $Q$ and dimension $D$ — enabling parameter sharing across inputs of arbitrary lengths and ensuring computational stability.

\subsection{Multi-granularity Hierarchical Sparse Encoding}
\textbf{Multi-granularity Segmentation.} In the multi-granularity encoding module, each channel of the input is independently processed to capture  intra-channel distinctive features. 
To capture intra-channel multi-scale temporal patterns, following \cite{wang2024medformer, zhang2024multi}, we partition each channel into multi-granularity segments using varying window sizes $\mathcal{S}=\{s_1, s_2, \ldots, s_G\}$ and $|\mathcal{S}|=G$. Each granularity $s_i$ generates a sequence of non-overlapping patches $\{p_1^{(i)}, p_2^{(i)}, \ldots\}$, where $p_j^{(i)} \in \mathbb{R}^{s_i}$ represents the $j$-th patch of granularity $i$.  The number of patches is $L_i=\lceil L / s_i\rceil$ with zero padding to ensure divisibility. These patches are projected into a unified latent space of dimension $D$ via linear transformations to obtain the patch embedding sequence $\mathbf{P}_i = [\hat{p}_1^{(i)}, \hat{p}_2^{(i)}, \ldots, \hat{p}_{L_i}^{(i)}] \in \mathbb{R}^{L_i \times D}$.
The embeddings of all granularities undergo hierarchical sparse encoding to model both intra-granularity and inter-granularity temporal dependencies.

\textbf{Intra-Granularity Hierarchical Sparse Encoding.} Each granularity is first processed independently to capture granularity-specific temporal dynamics.
Inspired by the human cognitive process of analyzing time series signals (e.g., ECG waveforms) \cite{wagenmakers2005human}—which involves iteratively filtering noise, aggregating local patterns, and distilling global semantics—we integrate $K$ TSDA blocks for intra-granularity processing to enable hierarchical feature refinement. 
The forward process of $k$-th TSDA block is formulated as follows:
\begin{equation}
\mathbf{H}_k=\text{TSDA}_{k}(\mathbf{H}_{k-1}; \mathbf{\Theta_k}, O_k)
\end{equation}
where $\mathbf{H}_{k-1}$ is the input token sequence and $\mathbf{H}_{0}=\mathbf{P}_i$. $\mathbf{H}_{k} \in \mathbb{R}^{O_k  \times D}$ is the output token sequence.
$\mathbf{\Theta_k}$ denotes trainable parameters and $O_k$ is the critical hyperparameter controlling token compression  and $O_k<O_{k-1}$. The output of the whole hierarchical TSDA processing is denoted as $\mathbf{H}^{\text{intra}} \in \mathbb{R}^{O_K  \times D}= \mathbf{H}_{K}$, a granularity-wise representation for subsequent inter-granularity correlation modeling. 


\textbf{Inter-Granularity Sparse Encoding.} Intra-granularity encoding has learned diverse high-level temporal features in different granularities, which are subsequently concatenated into a token sequence, denoted as $\mathbf{H}_\mathcal{S}^\text{intra} \in \mathbb{R}^{(G\cdot O^{\text{K}} ) \times D}=[\mathbf{H}_{s_1}^{\text {intra}} ; \mathbf{H}_{s_2}^{\text {intra}} ; \cdots ; \mathbf{H}_{s_G}^{\text {intra}}]$. A single TSDA block then models inter-granularity relationships as follows:
\begin{equation}
\mathbf{H}^\text {inter} \in \mathbb{R}^{O^{\text{inter}} \times D} =\text{TSDA}(\mathbf{H}_\mathcal{S}^\text{intra}; \mathbf{\Theta}, O^\text{inter})
\end{equation}
where $O^{\text{inter}} \ll  G\cdot O^{\text{K}}$. The self-attention of TSDA block serves to establish a global context among the tokens of different granularities. This operation allows the model to understand the overall structure and interconnections among the different granularity tokens. Then the token-sparse attention compresses the information from the different granularities into a more manageable and focused representation to refine the information as well as reduce computation. Additionally, different datasets may favor distinct granularities, and the domain knowledge-based learnable queries can guide the selection of optimal granularities for each dataset to enhance the model's generalization.

\subsection{Cross-Channel Sparse Encoding}
The output of inter-granularity encoding, $\mathbf{H}^{\text{inter}}$, serves as high-level semantic tokens for each channel. By concatenating all channel representations, we obtain the channel embedding matrix $\mathbf{H}^{\text{C}}=[\mathbf{H}_{1}^{\text {inter}} ; \mathbf{H}_{2}^{\text {inter}} ; \cdots ; \mathbf{H}_{C}^{\text {inter}}] \in \mathbb{R}^{(C \cdot O^{\text{inter}})\times D}$, which is then passed through a TSDA block to model and enhance inter-channel dependencies as follows:
\begin{equation}
\mathbf{H}_{\text{C}}^{\text{self}} \in \mathbb{R}^{(C \cdot O^{\text{inter}})\times D} \leftarrow \operatorname{Attn}^{\text {self}}\left(\mathbf{H}^\text{C}, \mathbf{H}^\text{C}, \mathbf{H}^\text{C}\right)
\label{eq:self}
\end{equation}
\begin{equation}
\mathbf{H}_{\text{C}}^{\text {sparse}} \in \mathbb{R}^{U \times D} \leftarrow \operatorname{Attn}^{\text {sparse}}\left(\mathbf{Q}_{\text{C}}^{\text {aug}}, \mathbf{H}_{\text{C}}^{\text {self}}, \mathbf{H}_{\text{C}}^{\text {self}}\right)
\label{eq:cross}
\end{equation}
The first self-attention layer in Equation \ref{eq:self} computes dense pairwise correlations across all channels, establishing a global context that captures both complementary relationships (e.g., spatially distant EEG channels jointly detecting propagating epileptic spikes). Leveraging the comprehensive context from the previous layer, the second token-sparse attention layer in Equation \ref{eq:cross} distills the $C$ channel tokens into $U$ ($U < C \cdot O^{\text{inter}}$) task-specific prototypes through learnable, domain-informed queries $\mathbf{Q}_{\text{C}}^{\text{aug}}$, aiming to filtering out irrelevant noise (e.g., overlapping functionalities among biosensors).  The output of TSDA block is flattened and projected into a D-dimensional space to generate the final temporal embedding:
$\mathbf{h}_i^{(x)} \in \mathbb{R}^{D}=\text{MLP}\left(\text{Flatten}\left(\mathbf{H}_C^{\mathrm{sparse}} \right)\right)$.
Note that this module's trainable parameters are independent of channel number $C$, allowing deployment across heterogeneous datasets with varying channel counts (e.g., $6$-channel ICU monitors vs. $12$-channel wearable arrays) without architectural adaptation. 
This refinement amplifies critical channel interactions while suppressing noise.


\subsection{Adaptive Label Encoder}
Traditional classification models rely on one-hot embedding for label representation, struggling to adapt to heterogeneous label spaces or generalize to unseen classes, limiting their cross-dataset transferability.
Recent advances attempt to mitigate this challenge: ZeroG \cite{li2024zerog} constructs a unified cross-dataset label space via pre-trained language model (LM) for graph classification. UniTS \cite{gao2024units} introduces trainable CLS tokens as label embeddings to support different time series classification task adaptation. Akata et al. \cite{7293699} utilize attribute embeddings as priors and update label embeddings for image classification through labeled training data. Inspired by these works, we propose an adaptive label encoder designed to enhance the model's cross-dataset transferability and generalization capabilities. A subsequent learnable projector dynamically refines the label embeddings, mapping them to a unified D-dimensional space shared with the time series embeddings. The formula is as follows:
\begin{equation}
\mathbf{h}_i^{(y)} \in \mathbb{R}^{D} =\mathbf{W}_1 \cdot (\text{ReLU}(\mathbf{W}_2 \cdot f_{\mathrm{LM}}(\mathcal{T}_{y_i}) + b))
\end{equation}
where $f_{\mathrm{LM}}$ refers to a frozen language model, and $T_{y_i}$ denotes the textual description of label $y_i$. $\mathbf{h}_i^{(y)}$ represents the adaptive label embedding.

\textbf{Loss Function:} A cross-entropy loss is utilized for training, formulated as follows:
\begin{equation}
\mathcal{L}=-\sum_{i=1}^N \log \frac{\exp \left(\operatorname{sim}\left(\mathbf{h}_i^{(x)}, \mathbf{h}_i^{(y)}\right)\right)}{\sum_{j=1}^M \exp \left(\operatorname{sim}\left(\mathbf{h}_i^{(x)}, \mathbf{h}_j^{(y)}\right)\right)}
\end{equation}
where sim(·) is a function to measure the similarity between temporal embedding and class embedding. During the inference stage, the class with the highest similarity score is predicted as label of the medical signal, formalized as:
\begin{equation}
y^{\prime}_{i}=\operatorname{argmax}_j\left(\operatorname{sim}\left(\mathbf{h}_i^{(x)}, \mathbf{h}_j^{(y)}\right) \mid j \in\{1, \ldots, M\}\right)
\end{equation}
where $y^{\prime}_{i}$ is the predicted label for sample  $\mathbf{X}_i$. We employ the dot product as the function sim(·).

\input{./Tables/datasets.tex}

\section{Experiments}
We evaluate our model's efficacy and in-domain/cross-domain transferability on $7$ real-world datasets from $3$ medical domains with $13$ baselines. 


\subsection{Experimental Setup}
\subsubsection{Datasets.}
We select datasets from three medical domains. \textbf{(1)} \textbf{Alzheimer's Disease}: APAVA \cite{escudero2006analysis} and ADFTD \cite{miltiadous2023dataset} are two EEG datasets for Alzheimer's Disease classification.
\textbf{(2)} \textbf{Epilepsy}: TUSZ v1.5.2 \cite{shah2018temple} is a large-scale corpus of EEG signals for Epilepsy. It offers two label sets: a coarse-grained label set (2 Classes) that distinguishes between seizure and non-seizure signals, and a fine-grained label set (4 Classes) that categorizes seizures into four types. \textbf{(3)} \textbf{Heart Disease}: PTB \cite{physiobank2000physionet} and PTB-XL \cite{wagner2020ptb} are two large-scale ECG databases for heart disease diagnosis. PTB-XL also provides two label sets: PTB-XL (4 Classes) with coarse-grained labels, and PTB-XL (5 Classes) with fine-grained labels. Table \ref{tab:datasets} provides brief information about the processed datasets. For more details regarding \textbf{data characteristics (e.g.  text description of class names, class distributions), dataset URL, train-validation-test splits, as well as data preprocessing}, dataset description $\mathcal{T}^\text{data}$, please see \textbf{Appendix 1.1}.

\subsubsection{Baselines.} We compare our model with a diverse set of baselines categorized as follows: \textbf{(1) Non-Transformer Models}: 
DLinear \cite{zeng2023dlinear}, 
MultiRocket \cite{tan2022multirocket}, 
LightTS \cite{zhang2022lightts}, 
TimesNet \cite{wu2022timesnet}.
\textbf{(2) Non-Multi-granularity Transformer-based Models}:  
PatchTST \cite{nie2023time}, Autoformer \cite{wu2021autoformer}, Crossformer \cite{zhang2023crossformer}, ETSformer \cite{woo2022etsformer}, FEDformer \cite{zhou2022fedformer}, Informer \cite{zhou2021informer}.
\textbf{(3) Multi-granularity Transformer-based Models}:  
PathFormer \cite{chen2024pathformer}, Medformer \cite{wang2024medformer}, MTST \cite{zhang2024multi}. 
Further details about the baselines are provided in the \textbf{Appendix 1.2}.

\subsubsection{Implementation Details.}
Following \cite{wang2024medformer}, we macro-averaged F1, macro-averaged AUROC, macro-averaged AUPRC and accuracy as metrics. We save the model with the best F1 score on the validation set and evaluate it on the test set.
Due to data imbalance, we use the F1 score in the main paper to better reflect model performance; results of other metrics are in the \textbf{Appendix}.
We find a set of hyper-parameters that perform optimally for most datasets through fine-tuning: multi-granularity window $\mathcal{S}=[25, 50, 100, 150]$; TSDA blocks $K=3$ with hierarchical token list $\{O_1,\cdots,O_K\}=[128, 64, 32]$; inter-granularity encoding output tokens $O^{\text{inter}}=10$ and cross-channel encoding output tokens $U=5$;  hidden dimension $D=128$.  
We adopt ClinicalBERT \cite{wang2023optimized} as the frozen LM. The batch size is set to $128$ for the ADFTD and PTB datasets, while $32$ for the remaining datasets. We employ the AdamW optimizer and the Cosine scheduler for learning rate decay. The training session is conducted with ten random seeds ($41$–$50$) on fixed training, validation, and test sets to compute the mean and standard deviation of model performance. Each training process runs for up to $60$ epochs, with early stopping if there is no improvement in F1 score of validation set for $7$ consecutive epochs. All experiments are conducted using the PyTorch framework on NVIDIA A$6000$ ($48$GB) GPU.


\input{./Tables/supervised_f1.tex}
\input{./Tables/zeroshot_f1.tex}
\subsection{Supervised Learning}
Key findings in Table \ref{tab:main_result} include: (1) MultiRocket, DLinear, and LightTS perform poorly due to their simplified architectures, which struggle with complex temporal dependencies. (2) Transformer-based models generally surpass traditional methods, underscoring the effectiveness of self-attention mechanism. (3) Five suboptimal performances in the multi-granularity models highlight the effectiveness of the multi-granularity mechanism in leveraging the contributions of different granularities to decision-making. (4) MedSpaformer outperforms all datasets, demonstrating its strong generalization capability. By emphasizing useful multi-granularity tokens and progressively discarding redundant information, it extracts higher-level channel interactions, enhancing performance compared to Medformer.

\input{./Figures/tex/fewshot_main}
\input{./Figures/tex/visual}
\input{./Figures/tex/efficiency}

\subsection{Few-shot Learning}
Few-shot learning addresses the label scarcity challenge by transferring knowledge from source domain with ample labeled data to target domain with limited labels. In this section, we pre-train all models on the source dataset, then fine-tune them on the target dataset under \{5, 10, 20, 30, 40, 50\}-shot settings. Since our baselines have fixed input dimensions, direct transfer between heterogeneous datasets is infeasible. We select source-target dataset pair with the same input length and channel counts, namely PTB-XL (4-Classes) and PTB-XL (5-Classes), TUSZ (2-Classes) and TUSZ (4-Classes). After pre-training, we freeze the model backbone and train a task-specific classification head for fine-tuning. 
Figure \ref{fig:fewshot_main} shows that (1) The performance of nearly all the models increases when the shots increases. And our model has the best performance on almost all the shots, demonstrating its robustness in transferability. (2) The transformer-based models usually have better performance than non-transformer models, which is consistent with the supervised model performance. (3) The performance gap between our model and baselines is more pronounced in PTB-XL than in TUSZ, showing its superior few-shot learning capacity in PTB-XL.

\subsection{Zero-shot Learning}
To evaluate the zero-shot transferability of our model, we conduct in-domain and cross-domain experiments. In-domain experiments transfer knowledge between datasets within the same domain. For example, when the target dataset is APAVA, we use the remaining datasets in "Alzheimer’s Disease" domain, specifically ADFTD, as sources. In contrast, cross-domain experiments involve pre-training our model on all datasets of the source domain and evaluating it on the test dataset from the target domain. Since our baselines lack zero-shot capability, we provide few-shot and supervised learning results for comparison.
In Table \ref{tab:zero_shot}, (1) There are four best performances in in-domain experiments and only three in cross-domain experiments, indicating that in-domain transfer exhibits stronger zero-shot performance than cross-domain. (2) Our model's zero-shot performance excels DLinear under 50 shots. (3) Our model's best zero-shot performance on APAVA, ADFTD, and PTB-XL (4-Classes) surpasses DLinear under supervised learning, likely due to the extensive pre-training data that captures a wider range of temporal patterns.

\input{./Tables/ablation_f1_main.tex}
\subsection{More Experiments}
\textbf{Ablation Study.}
To assess the impact of critical modules in our model, we perform ablation studies in four configurations. "W/O Multi-Granularity" uses single-granularity $\{25
\}$ to replace multi-granularity. "W/O Channel Attention" replaces cross-channel encoding with simple concatenation of all channel representations. "W/O Sparse Attention" substitutes token-sparse attention with self-attention. "W/O Label Encoder" uses one-hot encoding for ground truth.
In Table \ref{tab:ablation}, Multi-Granularity makes the most significant contribution, improving performance by approximately $7\%$ on average of mean performance. Sparse attention follows with an enhancement of about $6\%$. Channel attention contributes nearly $5\%$ as well, while the Label Encoder provides an additional improvement of around $2\%$. These results underscore the efficacy of our proposed mechanisms.



\textbf{Efficiency Analysis.}
In Figure \ref{fig:efficiency}, we compare the efficiency of our model against representative baselines on APAVA dataset, including the time required to train one epoch, F1 score, and trainable parameters.
FEDformer is the fastest and most lightweight model, ranking second in performance. Medformer is the second fastest and second smallest, with the third-best performance. MedSpaformer ranks third in training time and has 8.4 million parameters, smaller than TimesNet and PathFormer. While it sacrifices some training speed compared to FEDformer and Medformer, it achieves a significantly higher F1 score.
Overall, MedSpaformer presents a balanced option in the trade-off between efficiency and effectiveness. Its relatively high performance with a reasonable trainable parameter count and training time makes it a viable choice.

\textbf{Visualization.} To better visualize the learned representations from supervised learning, we use t-SNE \cite{maaten2008visualizing} to project the representations of representative models on the PTB-XL (5-Classes) dataset into a 2D space in Figure \ref{fig:visual}.
DLinear struggles to distinguish between different classes. FEDformer demonstrates better class separation, particularly for the dominant class, ST-T. MTST further improves by identifying the second-largest class, Myocardial Infarction, but fails in the small classes. In contrast, our model provides better discrimination among all classes. 


\textbf{Sensitivity Analysis.} Due to space constraints, we discuss the influence of critical hyper-parameters on our model's performance in \textbf{Appendix 1.6}. 
 
\section{Conclusion}
We propose MedSpaformer, a novel transformer-based model tailored for medical time series classification. 
By incorporating token-sparse dual-attention mechanism into multi-granularity cross-channel encoding, MedSpaformer effectively captures both intra- and inter-channel dependencies as well as multi-scale temporal patterns critical to medical signals. 
The combination of sparse encoding and an adaptive label encoder enables MedSpaformer to process heterogeneous datasets with few-shot and zero-shot transferability.  Extensive experiments validate its superiority, robustness, and adaptability to improve diagnostic performance in medical contexts. The \textbf{limitations, future work, and social impact} are discussed in \textbf{Appendix 2 \& 3}.

\bibliography{ref} 
\clearpage
\pagebreak
\end{document}

%% file: Figures/tex/model.tex
\begin{figure*}[!t] 
  \centering
  \includegraphics[width=0.99\textwidth]{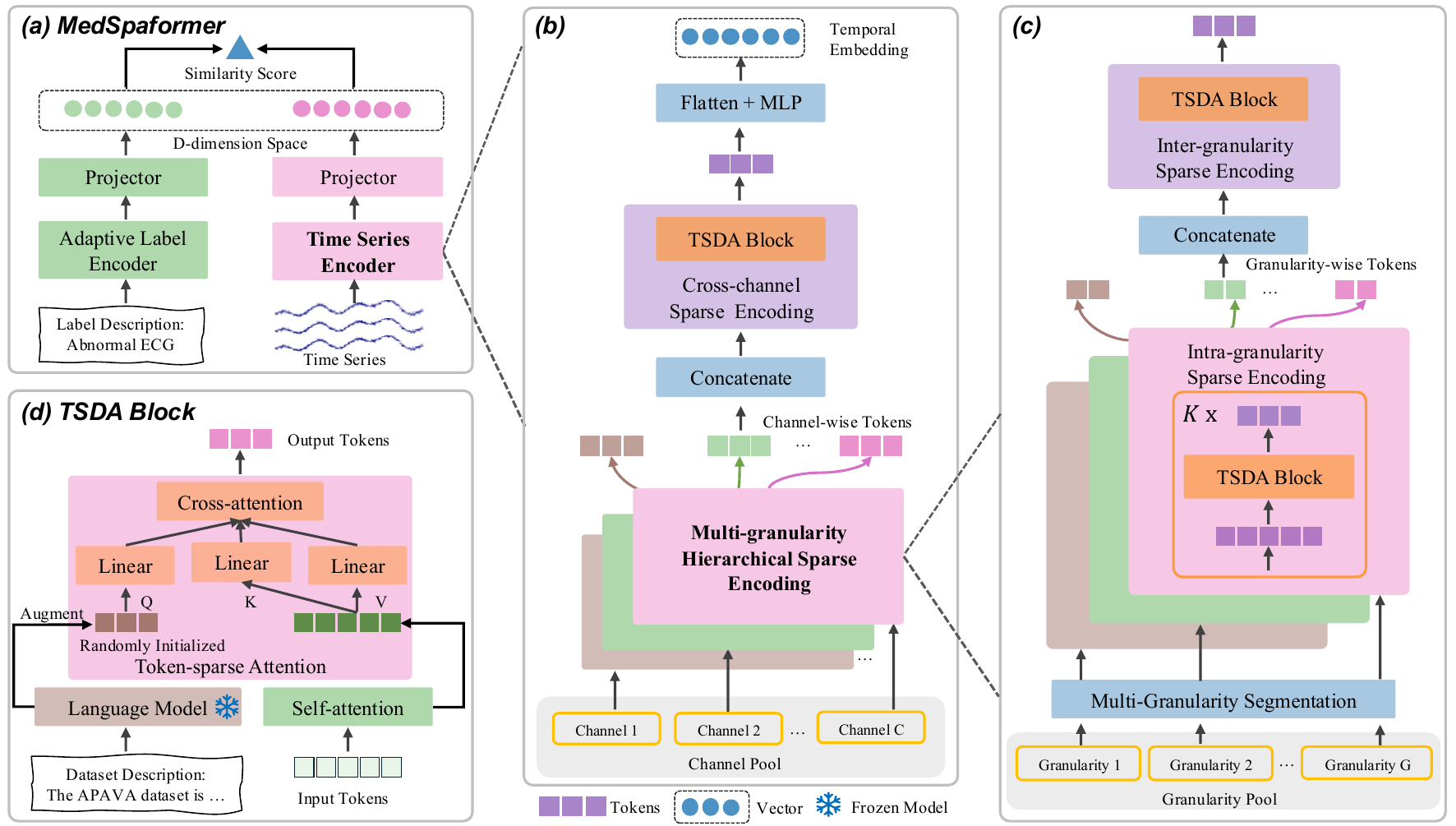}
  \caption{
(a) MedSpaformer consists of a time series encoder and a label encoder to map time series and labels into a unified space for optimization.
(b) shows the modeling of intra- and inter-channel correlations.
(c) illustrates multi-granularity encoding with TSDA blocks that capture intra- and inter-granularity dependencies.
(d) The token-sparse dual attention (TSDA) block combines self-attention to model global context and token-sparse attention to focus on informative local patterns.
  }
  \label{fig:model}
\end{figure*}


%% file: Tables/datasets.tex
\begin{table}[htb] 

\centering 
    \footnotesize
        \begin{threeparttable} 

\begin{tabular}{l|r|r|r}
\midrule \midrule
\textbf{Datasets}   & \textbf{\# Samples} & \textbf{\# Channels} & \textbf{\# Steps} \\ \midrule
APAVA (2-Classes)  & 5,967              & 16                  & 256              \\ \midrule
ADFTD (3-Classes)  & 69,752             & 19                  & 256              \\ \midrule
TUSZ (2-Classes)   & 22,040             & 19                  & 6,000            \\ \midrule
TUSZ (4-Classes)   & 2,891              & 19                  & 6,000            \\ \midrule
PTB (2-Classes)    & 64,356             & 15                  & 300              \\ \midrule
PTB-XL (4-Classes) & 17,110             & 12                  & 1,000            \\ \midrule
PTB-XL (5-Classes) & 17,110             & 12                  & 1,000            
            \\ 
            \midrule \midrule
            \end{tabular} 
            
        
        \end{threeparttable} 
        

\vspace{-1.5mm} 
\caption{Statistics of datasets.} 
\label{tab:datasets} 
\vspace{-4mm} 
\end{table} 

%% file: Tables/supervised_f1.tex

\begin{table*}[!t] 

\centering 

    
    \footnotesize
        \begin{threeparttable} 
\begin{tabular}{l|r|r|r|r|r|r|r}
\midrule \midrule


\textbf{\begin{tabular}[c]{@{}c@{}}Datasets\\ Model\end{tabular}} &
  \textbf{\begin{tabular}[c]{@{}c@{}}APAVA\\ (2-Classes)\end{tabular}} &
  \textbf{\begin{tabular}[c]{@{}c@{}}ADFTD\\ (3-Classes)\end{tabular}} &
  \textbf{\begin{tabular}[c]{@{}c@{}}TUSZ\\ (2-Classes)\end{tabular}} &
  \textbf{\begin{tabular}[c]{@{}c@{}}TUSZ\\ (4-Classes)\end{tabular}} &
  \textbf{\begin{tabular}[c]{@{}c@{}}PTB-XL\\ (4-Classes)\end{tabular}} &
  \textbf{\begin{tabular}[c]{@{}c@{}}PTB-XL\\ (5-Classes)\end{tabular}} &
  \textbf{\begin{tabular}[c]{@{}c@{}}PTB\\ (2-Classes)\end{tabular}} \\ \midrule
  
  
\textbf{MultiRocket} &
  0.511\text{\scriptsize ±0.015}&
  0.342\text{\scriptsize ±0.004}&
  0.629\text{\scriptsize ±0.029}&
  0.733\text{\scriptsize ±0.013}&
  0.224\text{\scriptsize ±0.018}&
  0.275\text{\scriptsize ±0.016}&
  0.572\text{\scriptsize ±0.006}\\
\textbf{Dlinear} &
  0.482\text{\scriptsize ±0.012}&
  0.295\text{\scriptsize ±0.002}&
  0.645\text{\scriptsize ±0.022}&
  0.736\text{\scriptsize ±0.016}&
  0.239\text{\scriptsize ±0.013}&
  0.251\text{\scriptsize ±0.001}&
  0.599\text{\scriptsize ±0.008}\\
\textbf{LightTS} &
  0.526\text{\scriptsize ±0.023}&
  0.378\text{\scriptsize ±0.021}&
  0.695\text{\scriptsize ±0.003}&
  0.843\text{\scriptsize ±0.014}&
  0.469\text{\scriptsize ±0.009}&
  0.431\text{\scriptsize ±0.011}&
  0.735\text{\scriptsize ±0.015}\\
\textbf{TimesNet} &
  0.703\text{\scriptsize ±0.019}&
  \underline{0.463\text{\scriptsize ±0.024}} &
  0.764\text{\scriptsize ±0.017}&
  0.849\text{\scriptsize ±0.012}&
  0.485\text{\scriptsize ±0.006}&
  0.526\text{\scriptsize ±0.022}&
  0.781\text{\scriptsize ±0.027}\\  \midrule
\textbf{PatchTST} &
  0.565\text{\scriptsize ±0.011}&
  0.453\text{\scriptsize ±0.015}&
  0.746\text{\scriptsize ±0.005}&
  0.855\text{\scriptsize ±0.022}&
  0.566\text{\scriptsize ±0.003}&
  0.509\text{\scriptsize ±0.024}&
  0.762\text{\scriptsize ±0.024}\\
\textbf{Autoformer} &
  0.715\text{\scriptsize ±0.021}&
  0.435\text{\scriptsize ±0.011}&
  0.725\text{\scriptsize ±0.018}&
  0.802\text{\scriptsize ±0.021}&
  0.432\text{\scriptsize ±0.019}&
  0.493\text{\scriptsize ±0.012}&
  0.635\text{\scriptsize ±0.014}\\
\textbf{Crossformer} &
  0.691\text{\scriptsize ±0.009}&
  0.428\text{\scriptsize ±0.014}&
  0.741\text{\scriptsize ±0.021}&
  0.837\text{\scriptsize ±0.011}&
  0.548\text{\scriptsize ±0.017}&
  0.485\text{\scriptsize ±0.007}&
  0.742\text{\scriptsize ±0.013}\\
\textbf{ETSformer} &
  0.652\text{\scriptsize ±0.022}&
  0.451\text{\scriptsize ±0.011}&
  0.811\text{\scriptsize ±0.024}&
  0.834\text{\scriptsize ±0.015}&
  0.508\text{\scriptsize ±0.013}&
  0.439\text{\scriptsize ±0.009}&
  0.803\text{\scriptsize ±0.021}\\
\textbf{FEDformer} &
  \underline{0.742\text{\scriptsize ±0.018}} &
  0.432\text{\scriptsize ±0.019}&
  0.718\text{\scriptsize ±0.012}&
  0.803\text{\scriptsize ±0.024}&
  0.528\text{\scriptsize ±0.023}&
  0.527\text{\scriptsize ±0.015}&
  0.686\text{\scriptsize ±0.012}\\
\textbf{Informer} &
  0.676\text{\scriptsize ±0.014}&
  0.461\text{\scriptsize ±0.008}&
  0.772\text{\scriptsize ±0.011}&
  0.848\text{\scriptsize ±0.003}&
  0.448\text{\scriptsize ±0.025}&
  0.463\text{\scriptsize ±0.006}&
  0.728\text{\scriptsize ±0.015}\\  \midrule
\textbf{PathFormer} &
  0.674\text{\scriptsize ±0.016}&
  0.415\text{\scriptsize ±0.009}&
  0.713\text{\scriptsize ±0.004}&
  0.794\text{\scriptsize ±0.022}&
  0.503\text{\scriptsize ±0.012}&
  0.482\text{\scriptsize ±0.002}&
  0.618\text{\scriptsize ±0.018}\\
\textbf{Medformer} &
  0.711\text{\scriptsize ±0.017}&
  0.459\text{\scriptsize ±0.013}&
  \underline{0.821\text{\scriptsize ±0.015}} &
  0.839\text{\scriptsize ±0.008}&
  \underline{0.575\text{\scriptsize ±0.014}} &
  0.519\text{\scriptsize ±0.005}&
  \underline{0.814\text{\scriptsize ±0.003}} \\
\textbf{MTST} &
  0.637\text{\scriptsize ±0.013}&
  0.427\text{\scriptsize ±0.012}&
  0.765\text{\scriptsize ±0.002}&
  \underline{0.855\text{\scriptsize ±0.009}} &
  0.546\text{\scriptsize ±0.011}&
  \underline{0.532\text{\scriptsize ±0.003}} &
  0.711\text{\scriptsize ±0.002}\\ \midrule
\textbf{MedSpaformer} &
  \textbf{0.821\text{\scriptsize ±0.014}} &
  \textbf{0.468\text{\scriptsize ±0.012}} &
  \textbf{0.852\text{\scriptsize ±0.007}} &
  \textbf{0.901\text{\scriptsize ±0.011}} &
  \textbf{0.583\text{\scriptsize ±0.014}} &
  \textbf{0.562\text{\scriptsize ±0.009}} &
  \textbf{0.843\text{\scriptsize ±0.014}}

            \\ 
            \midrule \midrule
            \end{tabular} 
            
        
        \end{threeparttable} 
        

\vspace{-1mm} 
\caption{Supervised Learning in F1 score and more analysis in other metrics are in \textbf{Appendix 1.3}. 
The best results are highlighted in red, while the second-best are in bold.} 
\label{tab:main_result} 
\vspace{-2mm} 
\end{table*} 

%% file: Tables/zeroshot_f1.tex
\begin{table*}[tb] 

\centering 

    \resizebox{\textwidth}{!}{ 
        \begin{threeparttable} 
        
            \begin{tabular}{c|l|cc|cc|ccc}
            \midrule \midrule
\multicolumn{2}{c|}{} &
  \multicolumn{7}{c}{\textbf{Test Datasets}} \\ \cmidrule{3-9} 
\multicolumn{2}{c|}{} &
  \multicolumn{2}{c|}{\textbf{Alzheimer's Disease}} &
  \multicolumn{2}{c|}{\textbf{Epilepsy}} &
  \multicolumn{3}{c}{\textbf{Heart Disease}} \\ 
\multicolumn{2}{c|}{\multirow{-3}{*}{\textbf{Zero-shot Experiments}}} & 

  \textbf{\begin{tabular}[c]{@{}c@{}}APAVA\\ (2-Classes)\end{tabular}} &
  \multicolumn{1}{c|}{\textbf{\begin{tabular}[c]{@{}c@{}}ADFTD\\ (4-Classes)\end{tabular}}} &
  \textbf{\begin{tabular}[c]{@{}c@{}}TUSZ\\ (2-Classes)\end{tabular}} &
  \multicolumn{1}{c|}{\textbf{\begin{tabular}[c]{@{}c@{}}TUSZ\\ (4-Classes)\end{tabular}}} &
  \textbf{\begin{tabular}[c]{@{}c@{}}PTB-XL\\ (4-Classes)\end{tabular}} &
  \textbf{\begin{tabular}[c]{@{}c@{}}PTB-XL\\ (5-Classes)\end{tabular}} &
  \textbf{\begin{tabular}[c]{@{}c@{}}PTB\\ (2-Classes)\end{tabular}} \\
  
  \midrule
\multicolumn{1}{c|}{} &

  \textbf{Alzheimer's Disease} &
  \cellcolor[HTML]{D0CECE}\textbf{0.533\text{\scriptsize±0.045}} &
  \cellcolor[HTML]{D0CECE}0.291\text{\scriptsize±0.062} &
  0.474\text{\scriptsize±0.048} &
  0.481\text{\scriptsize±0.030} &
  \textbf{0.285\text{\scriptsize±0.054}} &
  0.194\text{\scriptsize±0.025} &
  0.422\text{\scriptsize±0.099} \\
  
\multicolumn{1}{c|}{} &

  \textbf{Epilepsy} &

  0.407\text{\scriptsize±0.015} &
  0.273\text{\scriptsize±0.037} &
  \cellcolor[HTML]{D0CECE}0.470\text{\scriptsize±0.056} &
  \cellcolor[HTML]{D0CECE}\textbf{0.515\text{\scriptsize±0.032}} &
  0.238\text{\scriptsize±0.034} &
  0.173\text{\scriptsize±0.011} &
  0.381\text{\scriptsize±0.063} \\

\multicolumn{1}{c|}{\multirow{-3}{*}{\textbf{\begin{tabular}[c]{@{}c@{}}Pre-training\\      Domains\end{tabular} }}} &

  \textbf{Heart Disease} &
  0.413\text{\scriptsize±0.024} &
  \textbf{0.305\text{\scriptsize±0.049}} &
  \textbf{0.517\text{\scriptsize±0.068}} &
  0.506\text{\scriptsize±0.011} &
  \cellcolor[HTML]{D0CECE}0.271\text{\scriptsize±0.079} &
  \cellcolor[HTML]{D0CECE}\textbf{0.230\text{\scriptsize±0.034}} &
  \cellcolor[HTML]{D0CECE}\textbf{0.452\text{\scriptsize±0.055}}
  
\\ \midrule \midrule
\multicolumn{1}{c|}{} &
  \textbf{Dlinear (50-shot)} &
  N/A &
  N/A &
  0.434\text{\scriptsize±0.023} &
  0.478\text{\scriptsize±0.025} &
  0.187\text{\scriptsize±0.028} &
  0.163\text{\scriptsize±0.047} &
  N/A \\
\multicolumn{1}{c|}{} &
  \textbf{MedSpaformer (5-shot)} &
  N/A &
  N/A &
  0.546\text{\scriptsize±0.017} &
  0.577\text{\scriptsize±0.012} &
  0.253\text{\scriptsize±0.045} &
  0.259\text{\scriptsize±0.009} &
  N/A \\

\multirow{-3}{*}{\textbf{\begin{tabular}[c]{@{}c@{}}Suprevised/\\      Few-shot\\      Experiments\end{tabular}}} &

  \textbf{Dlinear (Supervised)} &
  0.482\text{\scriptsize±0.012} &
  0.295\text{\scriptsize±0.002} &
  0.645\text{\scriptsize±0.022} &
  0.736\text{\scriptsize±0.016} &
  0.239\text{\scriptsize±0.013} &
  0.251\text{\scriptsize±0.001} &
  0.599\text{\scriptsize±0.008}
  
            \\ 
            \midrule \midrule
            \end{tabular} 
            
        
        \end{threeparttable} 
        
        } 

\vspace{-1.5mm} 
\caption{Zero-shot Learning in F1 score and more results are in \textbf{Appendix 1.5}. : In-domain (gray background) versus cross-domain experiments, with comparisons to few-shot and supervised learning.
} 

\label{tab:zero_shot} 
\vspace{-1.5mm} 
\end{table*} 

%% file: Figures/tex/fewshot_main.tex
\begin{figure}[htb]
\centering

    \includegraphics[width=0.49\textwidth]{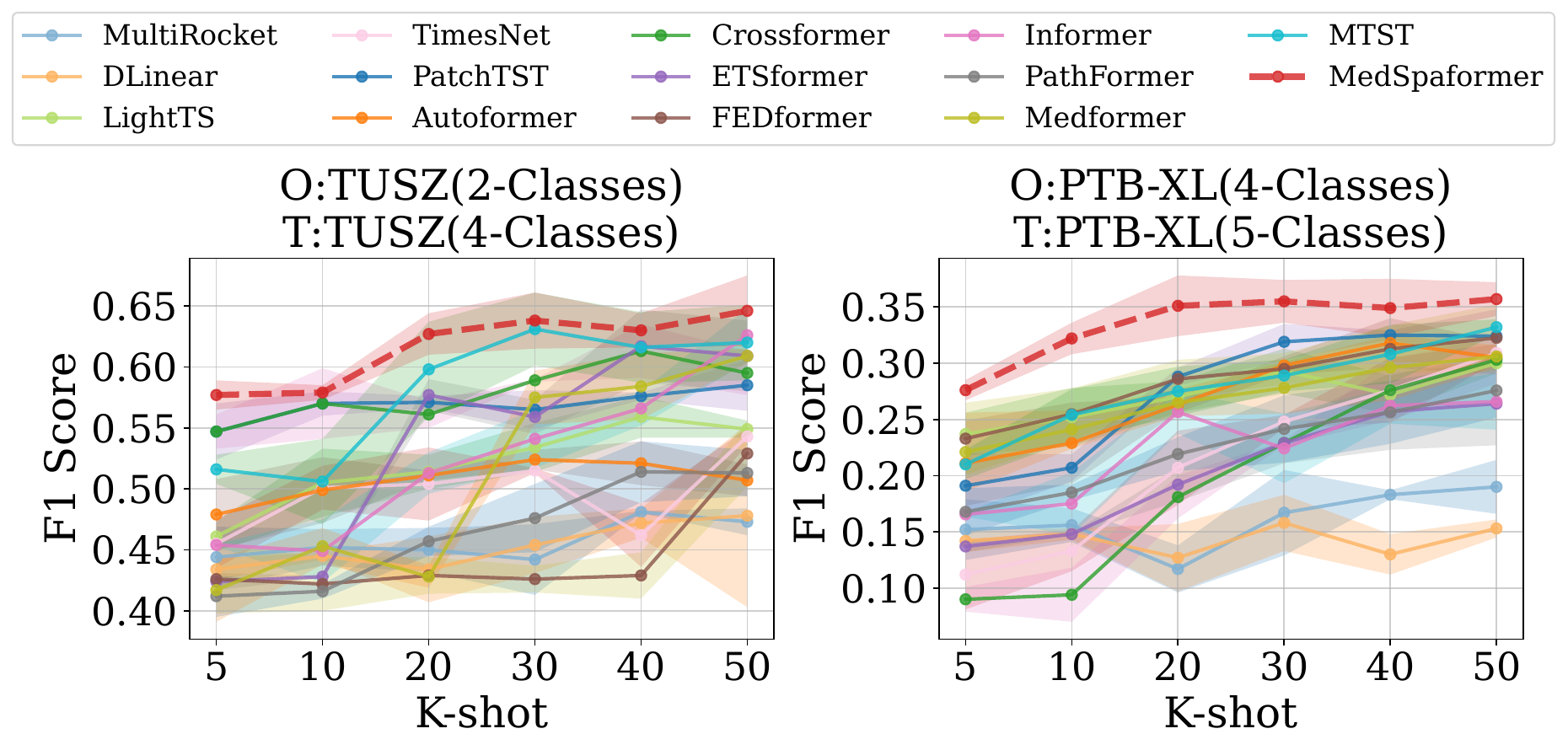}
      
    \vspace{-1.5mm} 
    \caption{Few-shot results under different shots on two experiments and more results are in \textbf{Appendix 1.4}.
    \textit{O} refers to source domain while \textit{T} refers to target domain.}
    
\label{fig:fewshot_main}
\vspace{-5mm} 
\end{figure}

%% file: Figures/tex/visual.tex
\begin{figure*}[htb]
  \begin{center}
    \includegraphics[width=0.95\textwidth]{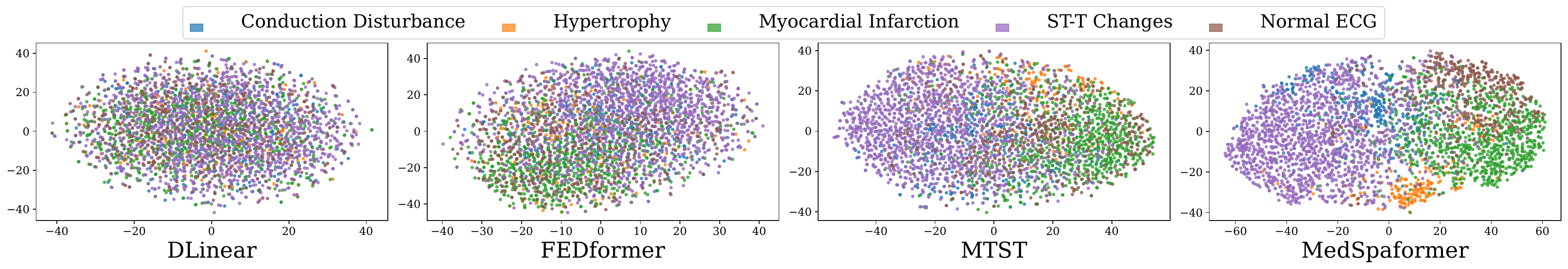} 
  \end{center}
  \vspace{-2.5mm}
  \caption{Embedding visualization of PTB-XL(5-Classes) on representative models, with colors indicating class labels.}
    \label{fig:visual}
    \vspace{-1.5mm}
\end{figure*}

%% file: Figures/tex/efficiency.tex
\begin{figure}[htb]
  \begin{center}
    \includegraphics[width=0.43\textwidth]{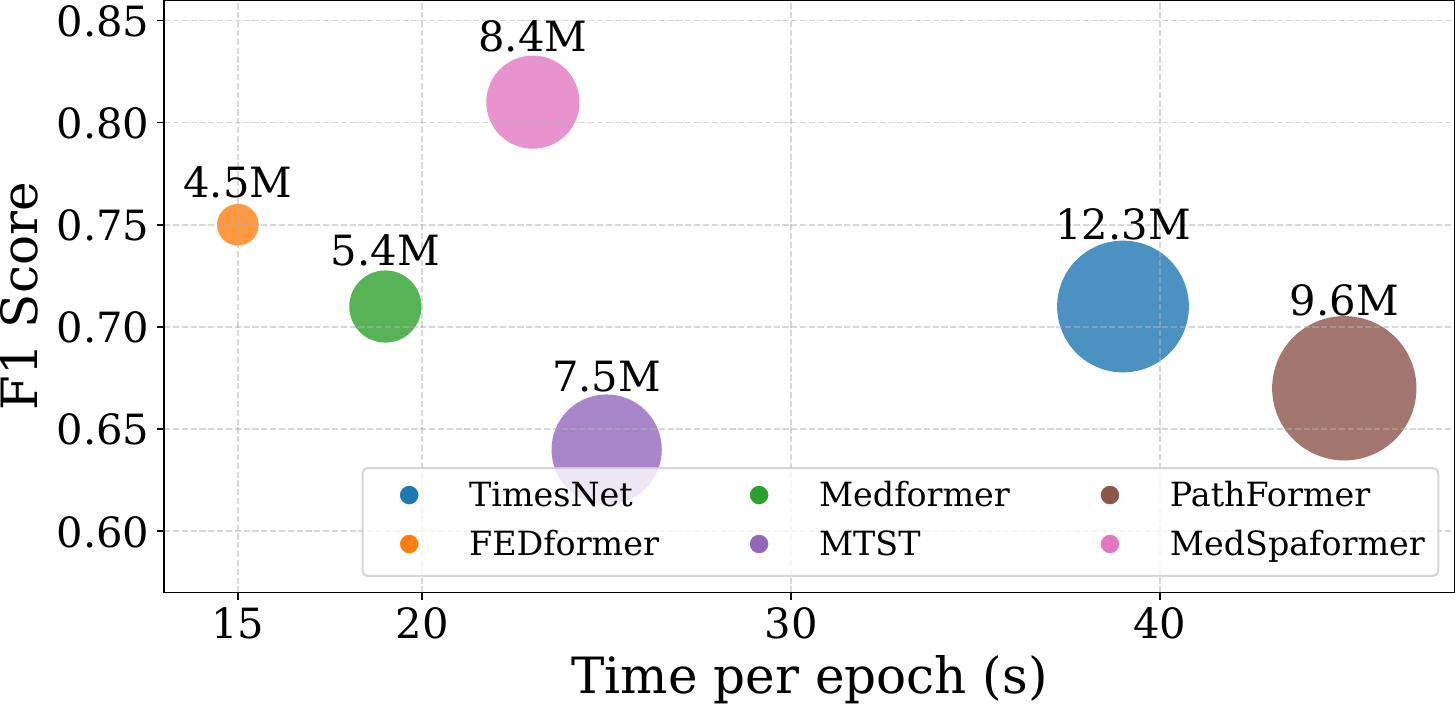} 
  \end{center}
  \vspace{-2mm}
  \caption{Efficiency comparison on the APAVA dataset (2 Classes). \textit{M} (million) serves as unit for trainable parameters.}
    \label{fig:efficiency}
    \vspace{-1.5mm}
\end{figure}

%% file: Tables/ablation_f1_main.tex
\begin{table}[htb] 

\centering 
    
    \resizebox{\columnwidth}{!}{ 
    
        \begin{threeparttable} 
            \begin{tabular}{l|c|c|c}
            \midrule \midrule
\multicolumn{1}{c|}{\textbf{Model}} &
  \multicolumn{1}{c|}{\textbf{\begin{tabular}[c]{@{}c@{}}APAVA\\      (2-Classes)\end{tabular}}} &
  \multicolumn{1}{c|}{\textbf{\begin{tabular}[c]{@{}c@{}}TUSZ\\      (2-Classes)\end{tabular}}} &
  \textbf{\begin{tabular}[c]{@{}c@{}}PTB\\      (2-Classes)\end{tabular}} \\

\midrule

\textbf{W/O Multi\_Granularity} & 0.727\text{\scriptsize ±0.012} & 0.771\text{\scriptsize ±0.003} & 0.753\text{\scriptsize ±0.008} \\
\textbf{W/O Channel Attention}  & 0.766\text{\scriptsize ±0.009} & 0.794\text{\scriptsize ±0.005} & 0.787\text{\scriptsize ±0.006} \\
\textbf{W/O Sparse Attention}   & 0.752\text{\scriptsize ±0.008} & 0.788\text{\scriptsize ±0.004} & 0.767\text{\scriptsize ±0.003} \\
\textbf{W/O Label Encoder}      & 0.796\text{\scriptsize ±0.019} & 0.828\text{\scriptsize ±0.007} & 0.820\text{\scriptsize ±0.015} \\

\textbf{MedSpaformer}           & \textbf{0.821\text{\scriptsize ±0.014}} & \textbf{0.852\text{\scriptsize ±0.007}} & \textbf{0.843\text{\scriptsize ±0.014}}

            \\ 
            \midrule \midrule
            \end{tabular} 
            
        
        \end{threeparttable} 
        
        } 

\vspace{-1.5mm} 
\caption{Ablation Study in F1 score on three datasets. More results are in \textbf{Appendix 1.6}.} 
\label{tab:ablation} 
\vspace{-3mm} 
\end{table} 